\documentclass{article}
\usepackage{spconf,amsmath,graphicx}

\usepackage{bm,amsmath,amssymb,mathrsfs,amsfonts}
\usepackage{caption}
\usepackage{scalefnt}
\usepackage{multirow}
\usepackage{here}
\usepackage{booktabs}
\usepackage{url}
\usepackage{enumitem}
\usepackage{type1cm}
\usepackage{cite}
\usepackage{arydshln}

\newcommand{\argmax}{\mathop{\rm argmax}\limits}
\newcommand{\ysrc}{Y^{\rm src}}
\newcommand{\ytgt}{Y}

\newcommand{\ytgtobs}{Y_{\rm obs}}

\newcommand{\ytgtmask}{Y_{\rm mask}}

\newcommand{\yhattgt}{\hat{Y}}

\newcommand{\yhattgtobs}{\hat{Y}_{\rm obs}}

\newcommand{\itert}{(t)}
\newcommand{\itertminusone}{(t-1)}

\newcommand{\ntgt}{N}
\newcommand{\nhatsrc}{\hat{N}_{\rm src}}
\newcommand{\nhattgt}{\hat{N}}

\AtBeginDocument{
  \abovedisplayskip     =0.5\abovedisplayskip
  \abovedisplayshortskip=0.5\abovedisplayshortskip
  \belowdisplayskip     =0.5\belowdisplayskip
  \belowdisplayshortskip=0.5\belowdisplayshortskip}

\usepackage{todonotes}

\title{Orthros: Non-autoregressive End-to-end Speech Translation\\with Dual-decoder}
\name{Hirofumi Inaguma\textsuperscript{\rm 1},
      Yosuke Higuchi\textsuperscript{\rm 2},
      Kevin Duh\textsuperscript{\rm 3},
      Tatsuya Kawahara\textsuperscript{\rm 1},
      Shinji Watanabe\textsuperscript{\rm 3}}
      
\address{\textsuperscript{\rm 1}Kyoto University, Japan
         \textsuperscript{\rm 2}Waseda University, Japan
         \textsuperscript{\rm 3}Johns Hopkins University, USA}
\begin{document}
\ninept
\maketitle
\begin{abstract}
Fast inference speed is an important goal towards real-world deployment of speech translation (ST) systems.
End-to-end (E2E) models based on the encoder-decoder architecture are more suitable for this goal than traditional cascaded systems, but their effectiveness regarding decoding speed has not been explored so far.
Inspired by recent progress in non-autoregressive (NAR) methods in text-based translation, which generates target tokens in parallel by eliminating conditional dependencies, we study the problem of NAR decoding for E2E-ST.
We propose a novel NAR E2E-ST framework, \textit{Orthros}, in which both NAR and autoregressive (AR) decoders are jointly trained on the shared speech encoder.
The latter is used for selecting better translation among various length candidates generated from the former, which dramatically improves the effectiveness of a large length beam with negligible overhead.
We further investigate effective length prediction methods from speech inputs and the impact of vocabulary sizes.
Experiments on four benchmarks show the effectiveness of the proposed method in improving inference speed while maintaining competitive translation quality compared to state-of-the-art AR E2E-ST systems. 
\end{abstract}
\begin{keywords}
End-to-end speech translation, non-autoregressive decoding, conditional masked language model
\end{keywords}

\vspace{-1mm}
\section{Introduction}\label{sec:intro}
\vspace{-1mm}
There is a growing interest in speech translation~\cite{ney1999speech} due to the increase in demand for international communications. 
The goal is to transform speech from one language to text in another language. 
Traditionally, the dominant approach is to cascade automatic speech recognition (ASR) and machine translation (MT) systems.
Thanks to recent progress in neural approaches, researchers are shifting to the development of end-to-end speech translation (E2E-ST) systems~\cite{listen_and_translate}, aiming to optimize a direct mapping from source speech to target text translation by bypassing ASR components.
The potential advantages of E2E modeling are (a) mitigation of error propagation from incorrect transcriptions, (b) low-latency inference, and (c) applications in endangered language documentation~\cite{unwritten_asru2017}.
However, most efforts have been devoted to investigating methods to improve translation \textit{quality} by the use of additional data~\cite{leveraging_weakly_supervised,pino2019harnessing,wang2020bridging,wang2020curriculum,inaguma19asru,di2019one,pino2020self,salesky-etal-2019-exploring,salesky2020phone}, better parameter initialization~\cite{berard2018end,bansal2018pre,indurthi2020end}, and improved training methods~\cite{weiss2017sequence,liu2019end}.

For applications of ST systems in lectures and dialogues, inference speed is also an essential factor from the user's perspective~\cite{niehues2016dynamic}.
Although E2E models are more suitable for small latency inference than cascaded models since the ASR decoder and the MT encoder processing can be skipped, their effectiveness regarding inference speed has not been well studied.
Moreover, incremental left-to-right token generation of \textit{autoregressive} (AR) E2E models increases computational complexity and, therefore, still suffer from slow inference.
To speed up inference while achieving comparable translation quality to AR models, \textit{non-autoregressive} (NAR) decoding has been investigated in text-based translation~\cite{gu2018non,lee2018deterministic,kaiser2018fast,libovicky2018end,ghazvininejad2019mask,gu2019levenshtein,stern2019insertion,ma2019flowseq,ghazvininejad2020aligned,saharia2020non}.
The NAR inference enables to generate tokens in parallel by eliminating conditional token dependencies.

Motivated by this, we propose a novel NAR framework, \textit{Orthros}, for the E2E-ST task.
Orthros has dual decoders on the shared speech encoder: NAR and AR decoders.
The NAR decoder is used for the fast token generation, and the AR decoder selects better translation among multiple length candidates during inference.
As the AR decoder can rescore all tokens in parallel and reuse encoder outputs, its overhead is minimal.
This architecture design is motivated by the difficulty of estimating a suitable target length given a speech in advance.
We adopt the conditional masked language model (CMLM)~\cite{ghazvininejad2019mask} for the NAR decoder, in which a subset of tokens are repeatedly updated by partial masking through constant iterations.
We also use semi-autoregressive training (SMART) to alleviate mismatches between training and testing conditions~\cite{ghazvininejad2020semi}.
However, any NAR decoder can be used in Orthros, conceptually.
Moreover, effective length prediction methods and the impact of vocabulary sizes are also studied.

Experiments on four benchmark corpora show that the proposed framework achieves comparable translation quality to state-of-the-art AR E2E- and cascaded-ST models with approximately 2.31$\times$ and 4.61$\times$ decoding speed-ups, respectively.
Interestingly, the best NAR model can even outperform AR models in terms of the BLEU score in some cases.
This work is the first study of NAR models for the E2E-ST task to the best of our knowledge.

\begin{table*}[t]
    \centering
    \footnotesize
    \caption{An example of Mask-Predict decoding on the Fisher-dev set. Highlighted tokens are masked for the next iteration.}\label{tab:result_case_study}
    \vspace{-3mm}
    \begingroup
    \begin{tabular}{l|l}\toprule
       Reference & but with that and and when we bought it i thought who's gonna put a you know a walkman or something and that's what i'm doing now \\

       $t=4$ & \colorbox{gray}{he came} with \colorbox{gray}{that} and when we bought \colorbox{gray}{it} i \colorbox{gray}{thought who going to put you know know ak or or something and} now i'm doing it \\
    
       $t=7$ & he came with that and when we bought it i thought who \colorbox{gray}{he going to put you know akman} or something and now i'm doing it \\
    
       $t=10$ & he came with that and when we bought it i thought who is going to put you know a walkman or something and now i'm doing it \\ 
       \bottomrule
    \end{tabular}
    \endgroup
    \vspace{-4mm}
\end{table*}

\vspace{-3mm}
\section{Background}\label{sec:background}

\vspace{-2.5mm}
\subsection{End-to-end Speech Translation (E2E-ST)}\label{ssec:e2e_st}
\vspace{-1mm}
E2E-ST is formulated as a direct mapping problem from input speech $X=(x_{1},\ldots,x_{U})$ in a source language to the target translation text $\ytgt=(y_{1}, \ldots, y_{\ntgt})$.
E2E models can be implemented with any encoder-decoder architectures, and we adopt the state-of-the-art Transformer model~\cite{vaswani2017attention}.
A conventional E2E-ST model is composed of a speech encoder and an \textit{autoregressive} (AR) translation decoder, which decomposes a probability distribution of $\ytgt$ into a chain of conditional probabilities from left to right as:
\vspace{-1.5mm}
\begin{eqnarray}
P(\ytgt|X) = \prod_{i=1}^{\ntgt} P_{\mathcal{\rm ar}}(y_{i}|y_{<i}, X). \label{eq:ar}
\end{eqnarray}
Parameters are optimized with a single translation objective $\mathcal{L}_{\mathcal{\rm ar}} = - \log P_{\mathcal{\rm ar}}(\ytgt|X)$ after pre-training with the ASR and MT tasks~\cite{berard2018end}.

\vspace{-2mm}
\subsection{Conditional masked language model (CMLM)}\label{ssec:nar}
\vspace{-1mm}
Since AR models generate target tokens incrementally during inference, they suffer from high inference latency and do not fully leverage the computational power of modern hardware such as GPU.
In order to tackle this problem, parallel sequence generation with \textit{non-autoregressive} (NAR) models have been investigated in a wide range of tasks such as text-to-speech synthesis~\cite{oord2018parallel,ren2019fastspeech}, machine translation~\cite{gu2018non}, and speech recognition~\cite{chen2019listen,chan2020imputer,higuchi2020mask,fujita2020insertion,bai2020listen,tian2020spike}.
NAR models factorize conditional probabilities in Eq.~\eqref{eq:ar} by assuming conditional independence for every target position.
However, iterative refinement methods generally achieve better quality than pure NAR methods at the cost of speed because of multiple forward passes~\cite{lee2018deterministic,ghazvininejad2019mask,guo-etal-2020-jointly}.
Among them, the conditional masked language model (CMLM)~\cite{ghazvininejad2019mask} is a natural choice because of its simplicity and good performance.
Moreover, we can flexibly trade the speed during inference by changing the number of iterations $T$.

In CMLM, the partial decoder inputs are masked by replacing with a unique token {\tt [MASK]} based on confidence.
Intermediate discrete variables at the $t$-th iteration $Y^{(t)}$ ($1 \leq t \leq T$) are iteratively refined given the rest observed tokens $ Y_{\rm obs}^{(t)} \subset Y^{(t-1)}$ as:
\vspace{-1mm}
\begin{eqnarray}
P(\ytgt|X) = \prod_{t=1}^{T} P_{\rm nar}(Y^{(t)}| Y_{\rm obs}^{(t)}, X).\label{eq:nar}
\end{eqnarray}
The target length distribution for $N$ is typically modeled by a linear classifier stacked on the encoder.

However, NAR models suffer from a \textit{multimodality} problem, where a distribution of multiple correct translations must be modeled given the same source sentence.
Recent studies reveal that sequence-level knowledge distillation~\cite{kim2016sequence} makes training data deterministic and mitigates this problem~\cite{gu2018non,zhou2019understanding,ren-etal-2020-study}.

\vspace{-1mm}
\section{Proposed method: Orthros}\label{sec:proposed_method}

\vspace{-2mm}
\subsection{Model architecture}\label{ssec:architecture}
\vspace{-1mm}
Typical text-based NAR models generate target sentences with multiple lengths in parallel to improve quality in a stochastic~\cite{gu2018non} or deterministic~\cite{ghazvininejad2019mask} way, followed by an optional rescoring step with a separate AR model~\cite{gu2018non}.
However, a spoken utterance consists of hundreds of acoustic frames even after downsampling, and its length varies significantly based on the speaking rate and silence duration.
Therefore, it is challenging to estimate the target length given a speech in advance accurately.
Moreover, extra computation and memory consumption for feature encoding with the separate AR model in rescoring are not negligible.
This motivated us to propose \textit{Orthros}, having dual decoders on top of the \textit{shared} encoder: an NAR decoder for fast token generation and an AR decoder for candidate selection from the NAR decoder.
This way, re-encoding speech frames is unnecessary, and the AR decoder greatly improves the effectiveness of using a large length beam.
The speech encoder is identical to that of AR models.
A length predictor and a CTC ASR layer for the auxiliary ASR task are also built on the same encoder.

Our NAR decoder is based on the conditional masked language model (CMLM)~\cite{ghazvininejad2019mask}.
One of the distinguished advantages over pure NAR models~\cite{gu2018non} is that CMLM removes the necessity of a copy of the source sentence to initialize decoder inputs.
This could be achieved by using a predicted transcription from the auxiliary ASR sub-module, but this contradicts a motivation to avoid ASR errors.

\vspace{-3mm}
\subsection{Inference}\label{ssec:inference}
\vspace{-1mm}
The inference of the CMLM is based on the \textit{Mask-Predict} algorithm~\cite{ghazvininejad2019mask}.
Let $T$ be the number of iterations, $\nhattgt$ be a predicted target sequence length, and $\hat{Y}_{\rm mask}^{\itert}$ and $\hat{Y}_{\rm obs}^{\itert}$ be masked and observed tokens in the prediction $\yhattgt^{(t-1)}$ ($|\yhattgt^{(t-1)}|=\nhattgt$) at the $t$-th iteration ($1 \leq t \leq T$), respectively.
At the initial iteration $t=0$, all tokens are initialized with {\tt [MASK]}. 
An example is shown in Table~\ref{tab:result_case_study}.

\vspace{-3mm}
\subsubsection{Mask-Predict}\label{ssec:mask_predict}
\vspace{-1mm}
The mask-predict algorithm performs two operations, \textit{mask} and \textit{predict}, at every iteration $t$.
In the \textit{mask} operation, given predicted tokens at the previous iteration, $\hat{Y}^{\itertminusone}$, we mask $k_{t}$ tokens having the lowest confidence scores, where $k_{t}$ is a linear decay function $k_{t}=\lfloor \nhattgt \cdot \frac{T-t}{t} \rfloor$.
In the \textit{predict} operation, we take the most probable token from a posterior probability distribution $P_{\rm cmlm}$ at every masked position $i$ and update $\hat{y}_{i}^{\itert} \in \hat{Y}_{\rm mask}^{\itert}$ as:
\begin{eqnarray}
\hat{y}_{i}^{\itert} = \argmax_{w_{i} \in V} P_{\rm cmlm}(w_{i}|\hat{Y}_{\rm obs}^{\itert}, X), \label{eq:y_sample}
\end{eqnarray}
where $V$ is the vocabulary.
When using SMART, described in Section~\ref{ssec:smart}, all tokens $\hat{y}_{i}^{\itert} \in \hat{Y}^{\itert}$ are updated in Eq.~\eqref{eq:y_sample} if they differ from those at the previous iteration.
Furthermore, we generate $l$ target sentences having different lengths in parallel and select the most probable candidate by calculating the average log probability over all tokens at the last iteration: $\frac{1}{\nhattgt}\sum_{i} \log P_{i, {\rm cmlm}}^{(T)}$.

For target length prediction, we sample top-$l$ $(l \geq 1)$ length candidates from a linear classifier conditioned on time-averaged encoder outputs.
We also study a simple scaling method using CTC outputs used for the auxiliary ASR task.

\vspace{-3mm}
\subsubsection{Candidate selection with AR decoder}
\vspace{-1mm}
Using multiple length candidates is effective for improving quality, but it is sub-optimal to directly use sequence-level scores from $P_{i, {\rm cmlm}}^{(T)}$ in Eq.~\eqref{eq:y_sample} because they are stale~\cite{ghazvininejad2019mask}.
Therefore, we propose to select the most probable translation among $l$ candidates after the last iteration by using log probability scores from the AR decoder averaged over all tokens.
Note that we do not use scores from the NAR decoder here.
Since the AR decoder can rescore all tokens in a candidate in parallel, it can still maintain the advantage of parallelism in self-attention.

\vspace{-3mm}
\subsection{Training}\label{ssec:training}
\vspace{-1mm}
The training objective of the CMLM can be formulated as follows:
\begin{eqnarray}
\mathcal{L}_{\rm cmlm} &=& - \sum_{y \in \ytgtmask} \log P_{\rm cmlm}(y|\ytgtobs, X), \label{eq:objective_v1}
\end{eqnarray}
where $\ytgtmask \subset Y$ and $\ytgtobs = Y \setminus \ytgtmask$.
We sample the number of masked tokens from a uniform distribution, $\mathcal{U}(1, \ntgt)$, following~\cite{ghazvininejad2019mask}.

\begin{table*}[t]
    \centering
    \begingroup
    \footnotesize
    \caption{BLEU scores of AR and NAR methods on the tst-COMMON sets of Must-C (En$\to$De and En$\to$Fr), Fisher-test (Fsh-test) and CallHome-evltest (CH-evltest) sets of Fisher-CallHome Spanish (Es$\to$En), and the test set of Libri-trans (En$\to$Fr). Seq-KD represents sequence-level knowledge distillation. Latency is measured as average decoding time per sentence on Must-C En$\to$De, with batch size 1.}\label{tab:result_main}
    \vspace{-3mm}
    \begin{tabular}{c|c|c|lccccccc}\toprule
      \multicolumn{2}{c}{} & \multicolumn{1}{c}{\multirow{4}{*}{ID}} & \multirow{4}{*}{Model} & \multicolumn{5}{c}{BLEU} & \multirow{4}{*}{\shortstack{Latency\\(ms)}} & \multirow{4}{*}{Speedup} \\ \cmidrule(r){5-9}
      
      \multicolumn{2}{c}{} & \multicolumn{1}{c}{} & \multicolumn{1}{c}{} & \multicolumn{2}{c}{Must-C} & \multicolumn{2}{c}{Fisher-CallHome} & \multirow{2}{*}{\shortstack{Libri-\\trans}} &  &  \\ \cmidrule(r){5-6} \cmidrule(r){7-8}

      \multicolumn{2}{c}{} & \multicolumn{1}{c}{} & \multicolumn{1}{c}{} & De & Fr & Fsh-test & CH-evltest & \\ \hline
     
        \multirow{15}{*}{\shortstack{E2E}} 
            & \multirow{4}{*}{AR} & \multirow{2}{*}{\texttt{A1}} & Transformer ($b=1$) & 21.54 & 32.26 & 48.38 & 18.07 & 16.52 & 175 & \phantom{0}1.54 $\times$ \\
            & & & Transformer ($b=4$) & \bf{23.12} & \bf{33.84} & \bf{48.49} & \bf{18.90} & \bf{16.84} & 271 & \phantom{0}1.00 $\times$ \\
            \cline{3-3} \cdashline{4-11}

            & & \multirow{2}{*}{\texttt{A2}} & Transformer + Seq-KD ($b=1$) & 23.88 & 33.92 & 50.34 & 19.09 & 15.91 & -- & -- \\
            & & & Transformer + Seq-KD ($b=4$) & \bf{24.43} & \bf{34.57} & \bf{50.32} & \bf{19.81} & 16.44 & -- & -- \\
            \cline{2-11}

            & \multirow{12}{*}{NAR} & \texttt{N1} & CTC ($b=1$) & 19.40 & 27.38 & 45.97 & 15.91 & 12.10 & \phantom{0}13 & 20.84 $\times$ \\ \cline{3-3} \cdashline{4-11}
            & & \multirow{4}{*}{\texttt{N2}} 
                & Orthros (CMLM, $T=4$) & 18.78 & 25.99 & 46.03 & 16.71 & 12.90 & -- & -- \\
            & & & Orthros (CMLM, $T=4$ +AR) & 19.62 & 27.77 & 47.80 & 18.28 & 13.69 & -- & -- \\
            & & & Orthros (CMLM, $T=10$) & 20.89 & 28.74 & 48.56 & 18.60 & 14.68 & -- & -- \\
            & & & Orthros (CMLM, $T=10$ +AR) & 21.79 & 30.31 & 49.98 & 19.71 & 15.43 & -- & -- \\ \cline{3-3} \cdashline{4-11}

            & & \multirow{4}{*}{\texttt{N3}} 
            & Orthros (SMART, $T=4$) & 20.03 & 27.22 & 45.89 & 17.39 & 14.17 & \phantom{0}46 & \phantom{0}5.89 $\times$ \\
            & & & Orthros (SMART, $T=4$ +AR) & 21.08 & 29.30 & 48.73 & 19.25 & 14.99 & \phantom{0}61 & \phantom{0}4.44 $\times$ \\
            & & & Orthros (SMART, $T=10$) & 21.25 & 29.31 & 47.09 & 18.25 & 15.11 & \phantom{0}99 & \phantom{0}2.73 $\times$ \\
            & & & Orthros (SMART, $T=10$ +AR) & \bf{22.27} & \bf{31.07} & \bf{50.07} & \bf{20.10} & \bf{16.08} & 111 & \phantom{0}2.44 $\times$ \\ \cline{3-3} \cdashline{4-11}
            & & \texttt{N4} & \ + BPE8k$\to$16k & \bf{22.88} & \bf{32.20} & \bf{50.18} & 19.88 & \bf{16.22} & 117 & \phantom{0}2.31 $\times$ \\ \cline{3-3} \cdashline{4-11}
            & & \multirow{2}{*}{\texttt{N5}} & \ \ + large (SMART, $T=4$ +AR, $l=7$) & 22.54 & 31.24 & -- & -- & -- & \phantom{0}59 & \phantom{0}4.59 $\times$ \\
            & & & \ \ + large (SMART, $T=10$ +AR, $l=7$) & \bf{23.92} & \bf{33.05} & -- & -- & -- & 113 & \phantom{0}2.39 $\times$ \\
            \hline 

        \multirow{2}{*}{\shortstack{Cascade}}
            & \multirow{2}{*}{AR} & \multirow{2}{*}{\texttt{A3}} & AR ASR ($b=1$) $\to$ AR MT ($b=1$) & 22.20 & 31.67 & 40.94 & 19.15 & 16.44 & 154$\to$166 & \phantom{0}0.84 $\times$ \\
            &  &  & AR ASR ($b=4$) $\to$ AR MT ($b=4$) & 23.30 & 33.40 & 42.05 & 19.77 & 16.52 & 333$\to$207 & \phantom{0}0.50 $\times$ \\
            \bottomrule
    \end{tabular}
    \endgroup
    \vspace{-3mm}
\end{table*}

\vspace{-2mm}
\subsubsection{Semi-autoregressive training (SMART)}\label{ssec:smart}
\vspace{-1mm}
To bridge the gap between training and test conditions in the CMLM, we adopt semi-autoregressive training (SMART)~\cite{ghazvininejad2020semi}.
SMART uses two forward passes to calculate the cross-entropy (CE) loss.
In the first pass, the CMLM generates predictions at all positions, $\yhattgt$, given partially-observed ground-truth tokens $\ytgtobs$ as in the original training process.
The gradient flow is truncated in the first pass~\cite{ghazvininejad2020semi}.
Then, a subset of tokens in $\yhattgt$ are masked again with a new mask.
The resulting observed tokens $\yhattgtobs$ are fed into the decoder as inputs in the second pass.
The CE loss is calculated with predictions at \textit{all} positions in the second pass, unlike the original training.

\vspace{-2mm}
\subsubsection{Total training objective}
\vspace{-1mm}
The speech encoder and all four branches (NAR/AR decoders, length predictor, and CTC ASR layer) are optimized jointly.
The total objective function is formulated as:
\begin{multline*}
\mathcal{L}_{\rm total} = (1- \lambda_{\rm asr}) \mathcal{L}_{\rm cmlm}(\ytgt|X) + \lambda_{\rm ar} \mathcal{L}_{\mathcal{\rm ar}}(\ytgt|X)  \\
+ \lambda_{\rm lp} \mathcal{L}_{\mathcal{\rm lp}}(\ntgt|X) + \lambda_{\rm asr} \mathcal{L}_{\rm asr}(\ysrc|X), \label{eq:total_loss}
\end{multline*}
where $\mathcal{L}_{\rm ar}$, $\mathcal{L}_{\rm lp}$, and $\mathcal{L}_{\rm asr}$ are losses in AR E2E-ST, length prediction, and ASR tasks, $\ysrc$ is the corresponding transcription, and $\lambda_{*}$ are the corresponding tunable hyperparameters.
We set ($\lambda_{\rm ar}$, $\lambda_{\rm lp}$, $\lambda_{\rm asr}$) to (0.3, 0.1, 0.3) throughout the experiments.

\vspace{-2mm}
\section{Experimental evaluation}\label{sec:experiments}
\vspace{-2mm}
\subsection{Datasets}\label{ssec:dataset}
\vspace{-1mm}
We used En-De (229k pairs, 408 hours) and En-Fr (275k pairs, 492 hours) language directions on Must-C~\cite{mustc}, Fisher-CallHome Spanish (Es-En, 138k pairs, 170 hours, hereafter Fisher-CH)~\cite{fisher_callhome}, and Libri-trans (En-Fr, 45k pairs, 100 hours)~\cite{libri_trans}.
All corpora contain a triplet of source speech and the corresponding transcription and translation, and we used the same preprocessing as~\cite{espnet-st}.
For Must-C, we report case-sensitive detokenized BLEU~\cite{bleu} on the {\tt tst-COMMON} set.
Non-verbal speech labels such as "\textit{(Applause)}" were removed during evaluation.
For Fisher-CH, we report case-insensitive detokenized BLEU on the Fisher-{\tt test} (four references) and CallHome-{\tt evltest} sets.
For Libri-trans, we report case-insensitive BLEU on the {\tt test} set.
We removed case information and all punctuation marks except for apostrophe in both transcriptions of all corpora and translations of Fisher-CH.

We extracted 80-channel log-mel filterbank coefficients with 3-dimensional pitch features using Kaldi~\cite{kaldi} as input speech features, which was augmented by a factor of 3 with speed perturbation~\cite{speed_perturbation} and SpecAugment~\cite{specaugment} to avoid overfitting.
All sentences were tokenized with the {\tt tokenizer.perl} script in Moses~\cite{moses}.
We built vocabularies based on byte pair encoding (BPE) algorithm~\cite{sennrich2015neural} implemented with Sentencepiece~\cite{sentencepiece}.
The joint source and target vocabularies were used in the ST/MT tasks, while the ASR vocabularies were constructed with the transcriptions only.
For Must-C, we used 5k and 8k vocabularies for ASR and E2E-ST/MT models, respectively.
For Fisher-CH and Libri-trans, we used 8k and 1k vocabularies for NAR E2E-ST models and the others, respectively.

\vspace{-3mm}
\subsection{Model configurations}\label{ssec:mode_configuration}
\vspace{-1mm}
We used the Transformer architecture implemented in ESPnet-ST~\cite{espnet-st} for all tasks.
All ASR and E2E-ST models consisted of stacked 12 encoder layers and 6 decoder layers.
Speech encoders had 2 CNN layers before the self-attention layers, which performed 4-fold downsampling.
The text encoders in the MT models consisted of 6 layers.
The dimension of self-attention layer $d_{\rm model}$ and feed-forward network $d_{\rm ff}$, and the number of heads $H$ were set to 256, 2048, and 4, respectively.
For the large model on Must-C, we set $d_{\rm model}=512$ and $H=8$.
The Adam optimizer~\cite{adam} with $\beta_{1}=0.9$, $\beta_{2}=0.98$, and $\epsilon=10^{-9}$ was used for training with Noam learning rate schedule~\cite{vaswani2017attention}.
Warmup steps and a learning rate constant were set to $25000$ and $5.0$, respectively.
A mini-batch was constructed with 32 utterances, and gradients were accumulated for 8 steps in NAR E2E-ST models.
The last 5 best checkpoints based on the validation performance were used for model averaging.
Following the standard practice in NAR models~\cite{gu2018non,ghazvininejad2019mask}, we used sequence-level knowledge distillation (Seq-KD)~\cite{kim2016sequence} with the corresponding AR Transformer MT model, except for Libri-trans.
During inference, we used a beam width $b \in \{1, 4\}$ for AR ASR/ST/MT models, and a length beam width $l=9$ for NAR models.
The language model was used for the ASR model on Libri-trans only.
Joint CTC/Attention decoding~\cite{hybrid_ctc_attention} was performed for ASR models.
Decoding time was measured with a batch size 1 on a single NVIDIA TITAN RTX GPU by averaging on five runs.
We initialized encoder parameters of E2E-ST models with those of the corresponding pre-trained ASR model and AR decoder parameters with those of the corresponding pre-trained AR MT model trained on the same triplets, respectively~\cite{berard2018end}.
However, NAR decoder parameters were initialized based on the weight initialization scheme in BERT~\cite{devlin2019bert,ghazvininejad2019mask}.

\begin{figure}[t]
  \vspace{-1mm}
  \centering
  \includegraphics[width=0.99\linewidth]{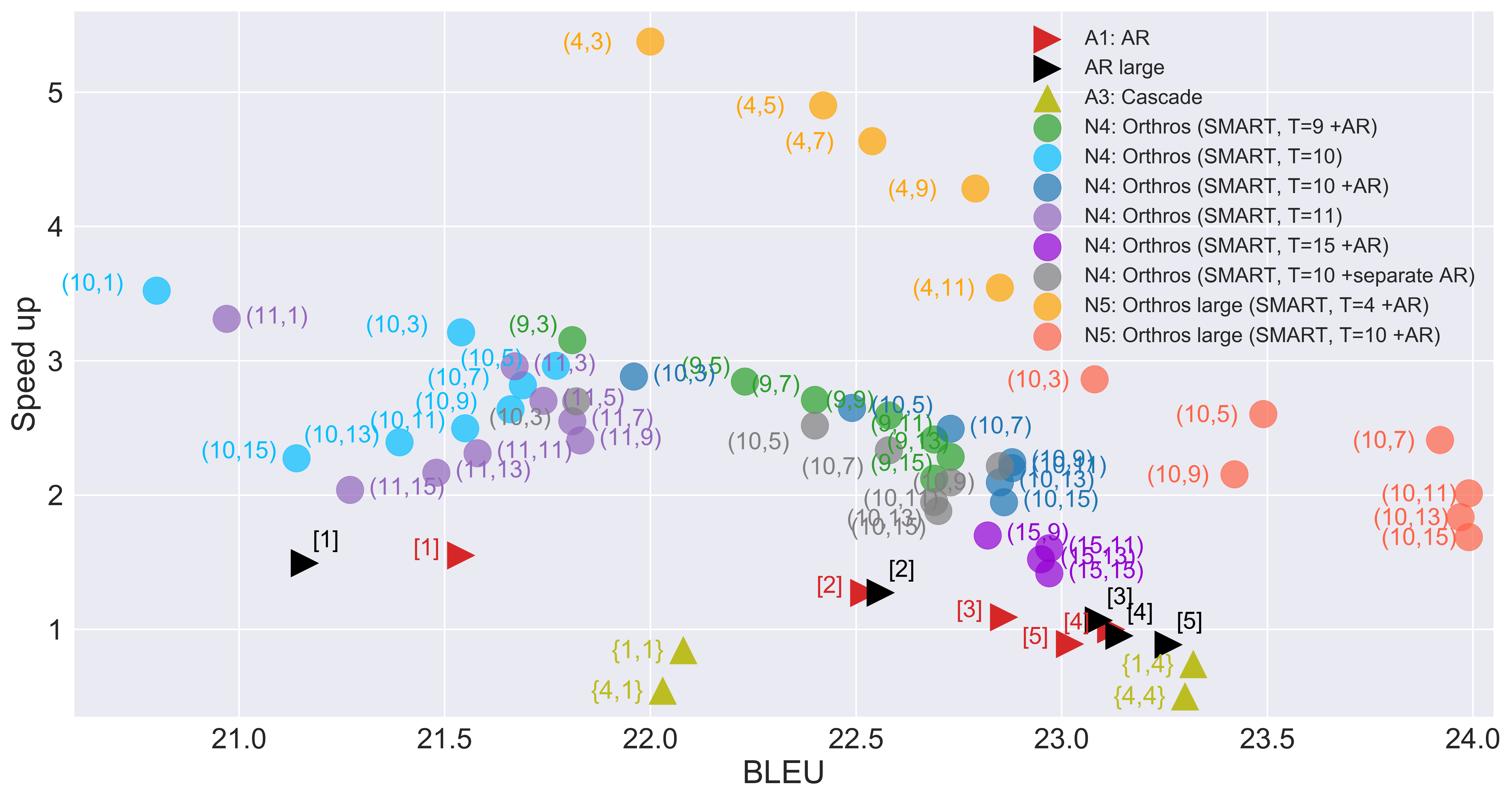}
  \vspace{-3mm}
  \caption{Trade-off between decoding speed-up and BLEU scores on the Must-C En-De tst-COMMON set. Parentheses, square brackets, and curly brackets represent ($T$, $l$), [$b$], and \{$b$ (ASR), $b$ (MT)\}, respectively.}
  \label{fig:2dgrid}
  \vspace{-1mm}
\end{figure}

\begin{table}[t]
    \vspace{-1mm}
    \centering
    \scriptsize
    \caption{Ablation study on the Fisher-dev set}\label{tab:result_ablation_study}
    \vspace{-3mm}
    \begingroup
    \begin{tabular}{l|cc|cc}\toprule
    \multirow{3}{*}{Model} & \multicolumn{4}{c}{BLEU} \\ \cline{2-5}
    & \multicolumn{2}{c|}{$T=4$} & \multicolumn{2}{c}{$T=10$} \\  \cline{2-5}
       & w/o AR & w/ AR & w/o AR & w/ AR \\ 
       \hline
      Orthros (\texttt{N3}) & \bf{45.76} & \bf{49.01} & 46.88 & \bf{50.28} \\
      \ - Seq-KD & 44.36 & 47.42 & 44.25 & 49.50 \\
      \ - AR decoder & 45.53 & N/A & \bf{46.94} & N/A \\
      \ + length prediction w/ CTC & 45.41 & 48.18 & 46.79 & 50.05 \\
      \bottomrule
    \end{tabular}
    \endgroup
    \vspace{-5mm}
\end{table}

\vspace{-2mm}
\subsection{Main results}
\vspace{-1mm}
The main results are shown in Table~\ref{tab:result_main}.
Iterative refinement based on CMLM (\texttt{N2}) significantly outperformed the pure NAR CTC model (\texttt{N1}) in translation quality.\footnote{CTC in the E2E-ST task has the speech encoder and the output classifier. It was optimized with a single CTC objective with a pair of ($X$, $\ytgt$). Since input speech lengths are generally longer than target sequence lengths in the E2E-ST task, we did not use the upsampling technique in~\cite{libovicky2018end}.}
Increasing the number of iterations $T$ was effective for improving quality at the cost of latency.
SMART also boosted the BLEU scores with no extra latency during inference, except for Fisher-CH (\texttt{N3}).
This is probably because sequence lengths in Fisher-CH are relatively short.
Candidate selection with the AR decoder greatly improved the quality with a negligible increase of latency, which corresponds to performing one more iteration.
We also found that NAR models prefer the large vocabulary size for better quality.
Using BPE16k, BLEU scores were improved while keeping latency (\texttt{N4}).
We note that the BPE size was tuned separately for AR and NAR models.
We will analyze this phenomenon in Section~\ref{ssec:vocab}.
Increasing the model capacity also improved the quality of NAR models while it did not hurt the speed so much when using GPU.
AR models did not benefit from the larger capacity on this corpus though not shown in the table (see Fig.~\ref{fig:2dgrid}).

For a comparison with AR models, Orthros achieved comparable quality to both strong AR E2E (\texttt{A1}, \texttt{A2}) and cascaded systems (\texttt{A3}) with smaller latency.
Interestingly, \texttt{N4} and \texttt{N5} even outperformed \texttt{A1} in quality by a large margin on Fisher-CH and Must-C En-De, respectively.
Seq-KD was very effective for AR models as well except for Libri-trans.
Although relative speed-ups are smaller than those in the MT literature~\cite{ghazvininejad2019mask,guo-etal-2020-jointly}, this is probably because we used the smaller vocabulary and the baseline AR models have much smaller latency (e.g., 607ms in \cite{guo-etal-2020-jointly} vs. 271ms in ours).

Fig.~\ref{fig:2dgrid} shows the trade-off between relative speed-ups and BLEU scores on the Must-C En-De tst-COMMON set.
Consistent with text-based CMLM models~\cite{ghazvininejad2019mask}, a large length beam width $l$ was not effective.
However, the proposed candidate selection significantly improved the performances with a larger $l$.
This way, a similar BLEU score can be achieved with a smaller iteration.
Moreover, Orthros (\texttt{N4}) can obtain the same BLEU as a baseline AR (\texttt{A1}) with more than 3 times speed-up for greedy decoding and 1.5 times for beam search.
The large Orthros (\texttt{N5}) achieved better BLEU scores than \texttt{N4} with similar latency and outperformed the AR models with beam search both in quality and latency.
Although the cascaded models showed reasonable BLEU scores, they were much slower than the E2E models.
We also compared AR models for candidate selection: the AR decoder on the unified encoder (proposal) vs. the separate AR encoder-decoder.
The unified encoder showed smaller latency with better quality.
We suspect that this is because sharing the encoder has a positive effect on candidate selection, or the AR decoder in Orthros was trained with Seq-KD.
We will analyze this in future work.
Although the overhead for additional speech encoding was relatively small here, this would be enlarged when using a more complicated encoder architecture.
One more advantage of Orthros is that the memory consumption for model parameters and encoder output caching is much smaller.

\vspace{-3mm}
\subsection{Ablation study}\label{ssec:ablation}
\vspace{-1mm}
To see individual contributions of the proposed techniques, we conducted the ablation study on the Fisher-dev set in Table~\ref{tab:result_ablation_study}.
Seq-KD was beneficial for boosting BLEU scores consistent with the NAR MT task~\cite{ghazvininejad2019mask}.
Joint training with the AR decoder did not hurt BLEU scores when candidate selection was not used.
For length prediction, we also investigated a simple approach by scaling the transcription length $\nhatsrc$, which was obtained from the CTC ASR layer with greedy decoding, by a constant value $\alpha$, i.e., $\nhattgt = \lfloor \alpha \nhatsrc \rfloor$.
Although this works as well, we needed to tune $\alpha$ on the dev set on each corpus, and therefore we adopted the classification approach.

\begin{figure}[t]
  \vspace{-1mm}
  \centering
  \includegraphics[width=0.99\linewidth]{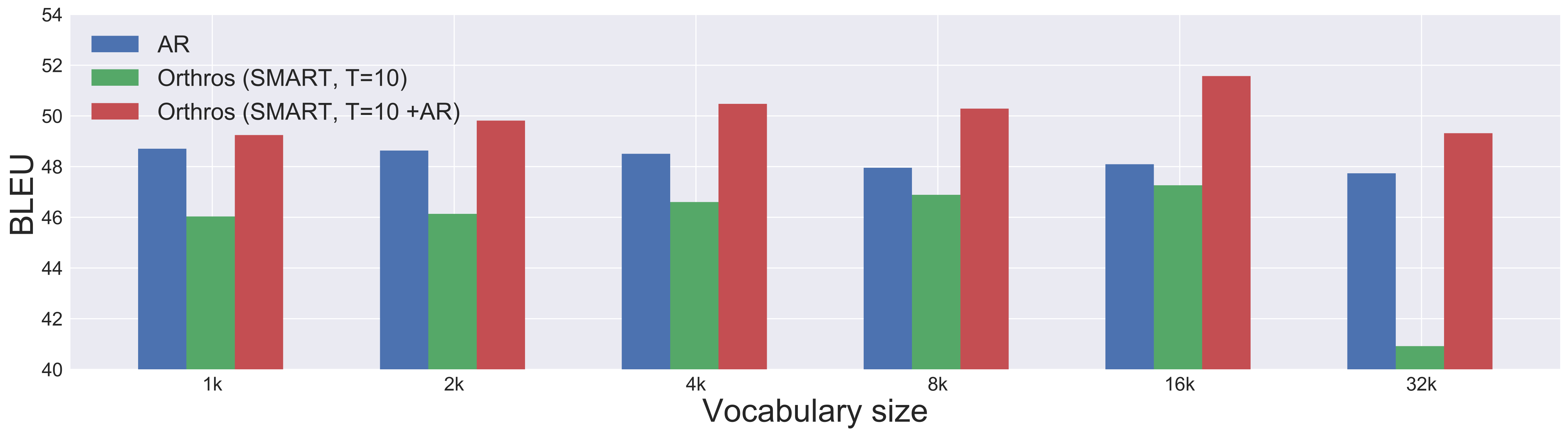}
  \vspace{-3mm}
  \caption{Impact of vocabulary size on the Fisher-dev set}
  \label{fig:bleu_vocab}
  \vspace{-5mm}
\end{figure}

\vspace{-3mm}
\subsection{Effect of vocabulary size}\label{ssec:vocab}
\vspace{-1mm}
Finally, we investigated the impact of vocabulary size.
Fig.~\ref{fig:bleu_vocab} shows BLEU scores of AR and NAR E2E models as a function of the vocabulary size on the Fisher-dev set.
We observed AR models have a peak around 1k BPE because the data size is relatively small (170-hours).
However, the performance of NAR models continued to improve according to the vocabulary size until 16k.
The candidate selection with the AR decoder was beneficial for all the vocabulary sizes, especially for 32k.
This is probably because misspelling was alleviated thanks to many \textit{complete} words in the large vocabulary, which had a complementary effect on the conditional independence assumption made in the NAR models.
We also observed similar trends in other corpora and CTC models.

\vspace{-2mm}
\section{Conclusion}\label{sec:conclusion}
\vspace{-1mm}
In this work, we proposed a unified NAR decoding framework to speed-up inference in the E2E-ST task, \textit{Orthros}, with NAR and AR decoders on the shared encoder.
Selecting the better candidate with the AR decoder greatly improved the effectiveness of a large length beam in the NAR decoder.
We also presented that using a large vocabulary and parameters is effective for NAR E2E-ST models.
The best NAR E2E model reached a level of state-of-the-art AR Transformer model in the BLEU score while reducing inference latency more than twice.

\fontsize{8.6pt}{0pt}\selectfont
\bibliographystyle{IEEEbib}
\bibliography{reference}

\end{document}